\documentclass[journal]{IEEEtran}
\usepackage{amsmath,amsfonts}
\usepackage{algorithmic}
\usepackage{algorithm}
\usepackage{array}
\usepackage[caption=false,font=normalsize,labelfont=sf,textfont=sf]{subfig}
\usepackage{textcomp}
\usepackage{stfloats}
\usepackage{url}
\usepackage{verbatim}
\usepackage{graphicx}
\usepackage{cite}

\usepackage{amsfonts,amssymb}
\usepackage{multirow}
\usepackage{stmaryrd}
\usepackage{color}

\usepackage[hidelinks]{hyperref}
\hypersetup
{
colorlinks = true,
linkcolor = red,
anchorcolor = black,
citecolor = green,
urlcolor = black
}

\begin{document}

\title{Quantity-Aware Coarse-to-Fine Correspondence for Image-to-Point Cloud Registration}

\author{Gongxin Yao, Yixin Xuan, Yiwei Chen and Yu Pan
\thanks{This work was supported by the National Natural Science Foundation of China under Grants No. U22A20102.}

\thanks{Gongxin Yao, Yixin Xuan, and Yu Pan are with the State Key Laboratory of Industrial Control Technology, Institute of Cyber-Systems and Control, College of Control Science and Engineering, Zhejiang University, Hangzhou 310027, China (e-mail:yaogongxin@zju.edu.cn; 22260380@zju.edu.cn; ypan@zju.edu.cn).}

\thanks{Yiwei Chen is with the Research Institute of Signals, Sensors and Systems, Heriot-Watt University, UK (e-mail: Yiwei.Chen@hw.ac.uk).}

\thanks{This paper has supplementary downloadable material available at
http://ieeexplore.ieee.org., provided by the author. The material includes a video showing point-to-pixel correspondences constructed by various methods in standard mp4 format. The total size of the video is 12.7 MB. Contact (yaogongxin@zju.edu.cn) for further questions about this work.}

}



\maketitle

\begin{abstract}
Image-to-point cloud registration aims to determine the relative camera pose between an RGB image and a reference point cloud, serving as a general solution for locating 3D objects from 2D observations. Matching individual points with pixels can be inherently ambiguous due to modality gaps. To address this challenge, we propose a framework to capture quantity-aware correspondences between local point sets and pixel patches and refine the results at both the point and pixel levels. This framework aligns the high-level semantics of point sets and pixel patches to improve the matching accuracy. On a coarse scale, the set-to-patch correspondence is expected to be influenced by the quantity of 3D points. To achieve this, a novel supervision strategy is proposed to adaptively quantify the degrees of correlation as continuous values. On a finer scale, point-to-pixel correspondences are refined from a smaller search space through a well-designed scheme, which incorporates both resampling and quantity-aware priors. Particularly, a confidence sorting strategy is proposed to proportionally select better correspondences at the final stage. Leveraging the advantages of high-quality correspondences, the problem is successfully resolved using an efficient \textit{Perspective-n-Point} solver within the framework of random sample consensus (RANSAC). Extensive experiments on the KITTI Odometry and NuScenes datasets demonstrate the superiority of our method over the state-of-the-art methods.
\end{abstract}


\section{Introduction}
\IEEEPARstart{I}{mage}-to-point cloud registration refers to the problem of estimating the camera pose (i.e., rotation and translation) of a query image with respect to the coordinate frame of a reference 3D point cloud. It has wide applications in various computer vision tasks, including but not limited to camera relocalization in visual Simultaneous Localization and Mapping (vSLAM) \cite{SLAMI, mur2015orb}, 3D reconstruction \cite{9950286, kerbl20233d}, and automatic camera extrinsic calibration for multi-sensor system with camera and LiDAR \cite{tsaregorodtsev2022extrinsic, liao2023se}. Generally, RGB images capture visual attributes like texture, color and luminosity through regular 2D grid data structures, while unordered and irregular point clouds solely provide rich 3D geometric information. Such modality gaps pose a significant challenge.

\begin{figure}[t]
\centering
\includegraphics[width=0.47\textwidth]{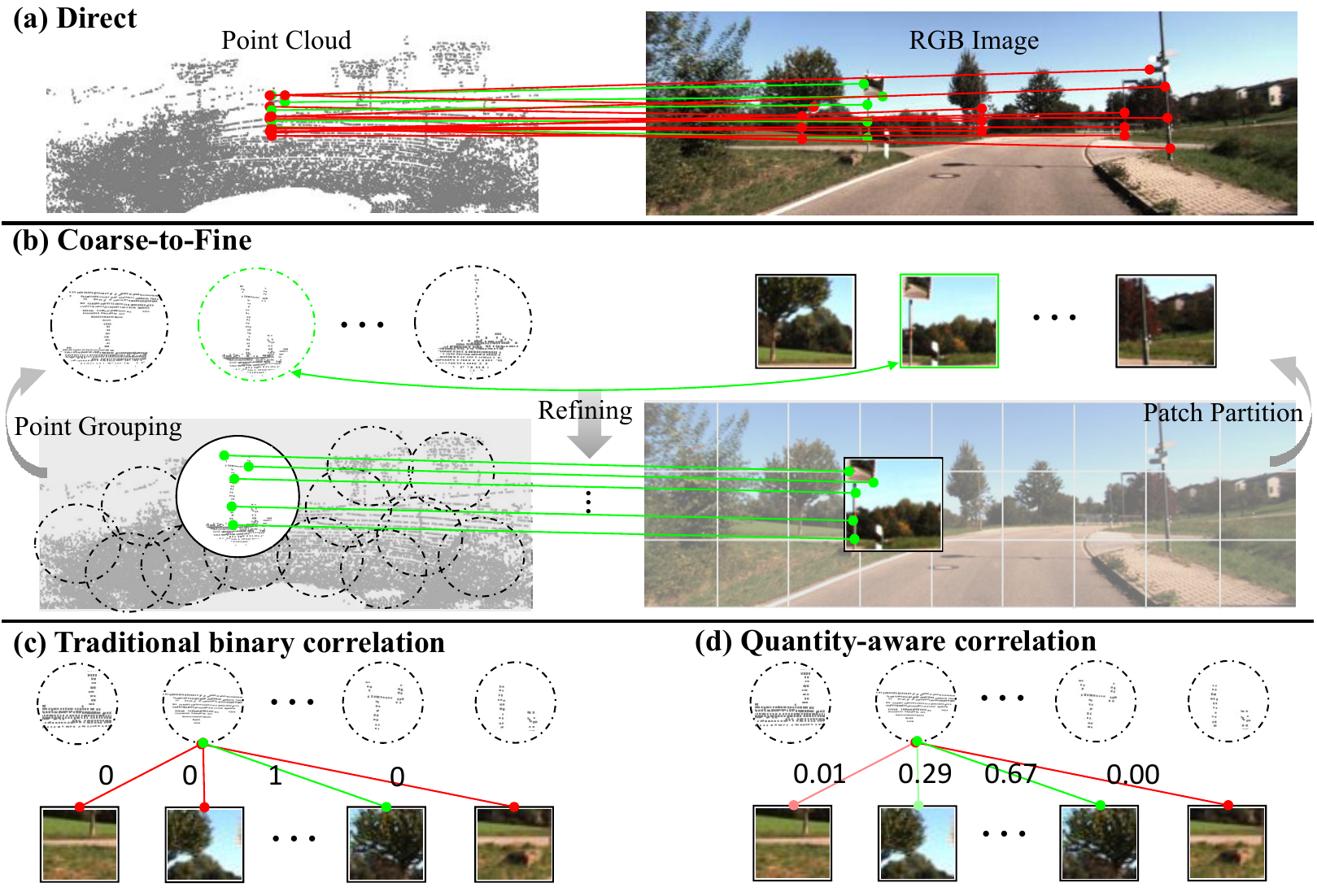} 
\caption{\textbf{Illustration of the coarse-to-fine mechanism and the set-to-patch correlation.} (a) Directly matching individual points with pixels is fraught with ambiguity due to the disparate visual attributes. (b) The global ambiguity between similar points and pixels can be resolved through the high-level semantics represented by local point sets and pixel patches. To supervise the matching between the sets and patches, the traditional binary correlation in (c) can be seen as a hard label (i.e., match-or-not), while our quantity-aware correlation in (d) can be seen as a soft label with richer information.}
\label{fig1}
\end{figure}

When the correspondences between 2D pixels and 3D points are available, the registration problem can be simplified to the traditional Perspective-n-Point (PnP) problem \cite{zheng2013revisiting}. In this context, both P3P \cite{2003Complete} and EPnP \cite{lepetit2009ep} algorithms have proven to be effective in estimating the camera pose. However, establishing the point-to-pixel correspondences is non-trivial and coupled with the unknown camera pose. In recent years, a promising paradigm has been proposed to match points and pixels in feature space via deep learning, thus providing a generalized correspondence extractor for the PnP algorithms. 2D3D-MatchNet \cite{feng20192d3d} adopts the hand-crafted SIFT \cite{lowe1999object} and ISS \cite{zhong2009intrinsic} operators to detect the interest pixels and points first, while using a metric learning framework \cite{ye2021deep} to learn discriminative descriptors for matching. DeepI2P \cite{li2021deepi2p} is a point-wise classification method that selects the points inside the camera frustum and solves the camera pose with inverse camera projection. Its multi-class classification variant matches each point with the pixels in a downsampled image. CorrI2P \cite{ren2022corri2p} predicts an overlap region and matches the descriptors of each pixel and point within it. However, these methods suffer from low inlier ratio because they solely operate on individual points and pixels, overlooking the high-level semantics conveyed by point sets and pixel patches.

The coarse-to-fine mechanism has been leveraged by our same-modal counterparts, i.e., image-to-image registration \cite{zhou2021patch2pix, sun2021loftr, mok2022affine} and point cloud registration \cite{yu2021cofinet, qin2022geometric}, which establish patch-to-patch and set-to-set correspondences first. Nevertheless, designing a coarse-to-fine pipeline that bridges the modality gaps, as well as addresses the unordered and irregular characteristics of point clouds, poses a non-trivial task. We attempt to migrate the coarse-to-fine mechanism to the field of image-to-point cloud registration, allowing us to match point sets with pixel patches based on their high-level semantics as illustrated in Fig. \ref{fig1} (b). In some concurrent works, GeoLoc \cite{Geoloc} and 2D3D-MATR \cite{li20232d3d} also based their work on similar ideas. However, their supervisions for matching the point sets and pixel patches are both binary values (i.e., match-or-not), which fails to reflect the misalignment between them. As illustrated in Fig. \ref{fig1} (c), a point set often overlaps with multiple pixel patches and should exhibit varying degrees of correlation.
 
In this study, we present a novel \textbf{C}oasre-to-\textbf{F}ine framework to learn quantity-aware correspondences for \textbf{I}mage-\textbf{to}-\textbf{P}oint \textbf{C}loud (CFI2P) registration. Following the design philosophy of Transformer \cite{vaswani2017attention} in point clouds \cite{yu2021pointr} and images \cite{dosovitskiyimage}, we translate both data into two sequences of local proxies that represent the local geometric features of point sets and pixel patches. Since the point sets have varying cardinalities, a redesigned attentive aggregation module with scattered indexing is proposed to generate proxies. On a coarse scale, the matching between local proxies is interpreted as an optimal transport problem \cite{sarlin2020superglue}, which is supervised by a correlation matrix between point sets and pixel patches. Considering the spatial misalignment, a novel quantity-aware strategy is proposed to quantify the degrees of correlation as continuous values in Fig. \ref{fig1} (d). It is achieved by calculating the proportions of 3D points from a bilateral perspective, i.e., one side is on the current point set, and the other side is on pixel patches. The calculation relies exclusively on the count of 3D points, without incorporating the quantity of any 2D pixels. This design grants our strategy a flexible adaptability to variations in point cloud density and addresses potential gaps in 2D-3D resolution. On a finer scale, we refine the set-to-patch correspondences to point and pixel levels through a well-designed scheme. It involves resampling, attentive learning and fine matching, where the latter two steps integrate the sampling priors. Positioned at the end of the fine matching, a novel confidence sorting strategy is also proposed to proportionally select better point-to-pixel correspondences based on the quantity-aware priors. Finally, all the results are collected to estimate the camera pose with a random sample consensus (RANSAC) based EPnP solver.

To summarize, our main contributions are as follows:
\begin{itemize}
    \item A novel coarse-to-fine correspondence learning framework for image-to-point cloud registration is proposed, emphasizing the learning of quantity-aware correspondences between point sets and pixel patches.
    \item A novel quantity-aware strategy to quantify the degrees of correlation between point sets and pixel patches as continuous values is introduced, which effectively captures spatial misalignment, offering richer supervision information. Additionally, it adapts to resolution gaps and varying point density for enhanced performance.
    \item A well-designed refining scheme with a confidence sorting strategy to establish high-quality point-to-pixel correspondences is proposed. This scheme fully leverages quantity-aware priors, effectively mitigating the negative effects of set-to-patch misalignment.
    \item Comprehensive experiments are conducted on the KITTI Odometry \cite{geiger2013vision} and NuScenes \cite{caesar2020nuscenes} datasets, demonstrating the state-of-the-art performance of our proposal.
\end{itemize}

The remainder of this paper is organized as follows. Section \ref{sec:related} briefly reviews some related works. Section \ref{sec:analysis} analyzes the challenges and presents our quantity-aware correlation strategy. Section \ref{sec:framework} introduces the workflow of our CFI2P framework in detail. Section \ref{sec:experiment} presents the experimental setting and results. Finally, Section \ref{sec:conclusion} concludes this study.

\section{Related Works}
\label{sec:related}
\subsection{Image-to-Image Registration}
Image-to-image registration \cite{ma2021image} is a multi-view geometric problem, which aims to estimating the relative pose between two images taken from different locations. Due to the lack of depth information, epipolar constraint \cite{diel2005epipolar, zhou2021patch2pix} is a classic method to describe the geometric relationship between the pixels in two different images. Besides, many algorithms are based on pixel-to-pixel feature matching. For example, FFT\cite{reddy1996fft}, SFT \cite{lowe1999object} and ORB \cite{rublee2011orb} are hand-crafted operators to extract local geometric descriptors. SuperGlue \cite{sarlin2020superglue}, LoFTR \cite{2021LoFTR}, ACVNet \cite{Xu_2022_CVPR}, AANet \cite{aanet2023}, TCDesc \cite{tcdesc2023} and Scale-Net \cite{calenet2023} are the representative learning-based methods.


\subsection{Point Cloud Registration}
Due to the inherent 3D geometric information, estimating the rigid transformation between two point clouds is relatively straightforward. ICP \cite{ICP1992} and its variants \cite{EVICP2001, yang2013go} can directly iterate on the 3D points to optimize the transformation, but they are sensitive to initialization. DCP \cite{wang2019deep} and PointNetLK \cite{aoki2019pointnetlk} are the seminal works that introduce deep learning to point cloud registration. After that, the recent works \cite{huang2021predator, lu2021hregnet, yu2021cofinet, qin2022geometric, ao2023buffer} adopted the architecture of KPConv \cite{thomas2019kpconv} to extract local features. Although KPConv requires complex data preprocessing, it supports to establish accurate correspondences from downsampled nodes to fine 3D points gradually. Beside, EDFNet \cite{zhang2022learning} learns the task-specific 3D point descriptor and SACF-Net \cite{sacfnet2023} is a correspondence filtering network to improve the quality of matching.

\subsection{Image-to-Point Cloud Registration}
Considering the modality gaps between images and point clouds, 2D3D-MatchNet \cite{feng20192d3d} detects sparse key pixels and points by SIFT \cite{lowe1999object} and ISS \cite{zhong2009intrinsic} first. The surrounding patches of these key pixels and points are fed into a pseudo-siamese neural network \cite{hoffer2015deep, xu2019deep} to learn point and pixel descriptors and establish 2D-3D correspondences. Camera pose estimation is then performed using a RANSAC-based EPnP \cite{lepetit2009ep, fischler1981random} solver. However, there is no guarantee that the extracted pixels and points are spatially correlated since SIFT and ISS have different feature extraction modes. DeeI2P \cite{li2021deepi2p} is a detection-free method that interprets the registration problem as binary classification, whereby the point cloud projected in the camera frustum is classified and then utilize inverse camera projection to estimate camera pose. CorrI2P \cite{ren2022corri2p} follows the same metric learning paradigm as 2D3D-MatchNet to learn descriptors for all pixels and points, while predicting a symmetric overlap region between images and point clouds. To establish point-to-pixel correspondences, each 3D point should search for its nearest 2D pixel in both the feature space and the overlap region. Some concurrent works, i.e., GeoLoc \cite{Geoloc} and 2D3D-MATR \cite{li20232d3d}, follow the coarse-to-fine mechanism to build point-to-pixel correspondences. GeoLoc is towards satellite RGB images and bird's-eye views of point clouds. 2D3D-MATR is a complex multi-scale framework. Although they are slightly different, their supervisions for matching the point sets and pixel patches are both binary values (i.e., match-or-not). It fails to adaptively reflect the degree of correlation, particularly in cases with varying point cloud density or when significant resolution gaps exist in representing the same object.

\section{Quantity-Aware Correlation}
\label{sec:analysis}
In this section, we first present the problem formulation of image-to-point cloud registration. Afterwards, we elaborate on the challenges in quantifying the degree of correlation between point sets and pixel patches, and then derive our quantity-aware strategy.

\subsection{Problem Formulation}
\label{sec:problem}
Given a query RGB image $ \textbf{I} \in \mathbb{R}^{H \times W \times 3} $ with a resolution of $H \! \times \! W$, and a reference point cloud $\textbf{P} \in \mathbb{R}^{N \times 3}$ with $N$ points $\textbf{p}_{i} \in \mathbb{R}^{3}$, our goal is to estimate a rigid transformation $\textbf{T} = [\textbf{R}|\textbf{t}] $ (i.e., the camera pose) with a rotation $\textbf{R} \in SO(3)$ and a translation $\textbf{t} \in \mathbb{R}^{3}$, which aligns the coordinate frames of the camera and point cloud. It is equivalent to minimizing the following objective function:
\begin{align}
    \label{eq1}
	\overline{\textbf{T}} &= \mathop{\mathrm{arg\,min}}\limits_{\textbf{T}} \sum\nolimits_{(\textbf{u}_{i},\textbf{p}_{i}) \in C^{*}} \Vert \textbf{u}_{i} - \mathcal{F}_{p}(\textbf{KT}\textbf{p}_{i})\Vert,
\end{align}
where $\Vert \cdot \Vert$ represents the Euclidean distance, $C^{*}$ is the set of ground-truth point-to-pixel correspondences, $\textbf{u}_{i} = [u_i, v_i]^{\mathsf{T}}$ is a 2D pixel coordinates, $\textbf{K} \in \mathbb{R}^{3 \times 3}$ is the camera intrinsic matrix, $\textbf{Tp}_{i}$ implicitly converts $\textbf{p}_{i}$ from inhomogeneous to homogeneous coordinates (i.e., $\textbf{p}_{i} \! \to \! [\textbf{p}_{i}, 1]^{\mathsf{T}}$) and then performs matrix product. $\mathcal{F}_{p}(\cdot)$ is a planar projection function:
\begin{align}
    \label{eq2}
	\mathcal{F}_{p}([x, y, z]^{\mathsf{T}}) = [u', v']^{\mathsf{T}} = [x/z, y/z]^{\mathsf{T}}.
\end{align} 

Following the paradigm of estimating the set $C^{*}$ and then solving the camera pose with RANSAC-based EPnP \cite{lepetit2009ep}, our primary focus is on the former. We attempt to match the higher-level semantic features between point sets and pixel patches, thereby mitigating the ambiguity of direct point-to-pixel matching. Denote the point sets as $\{G^{\textbf{P}}_{j}\}_{j=1}^{N_q}$ and the pixel patches as $\{\textbf{I}_i\}_{i=1}^{N_I}$. We should minimize the distance between $\textbf{I}_i$ and $G^{\textbf{P}}_{j}$ in feature space when they are correlated, and maximize it when they are not. Therefore, how to quantify such cross-modal correlation is the first core problem. It will serve as a supervision for the subsequent feature learning.

\begin{figure}[t]
\centering
\includegraphics[width=0.45\textwidth]{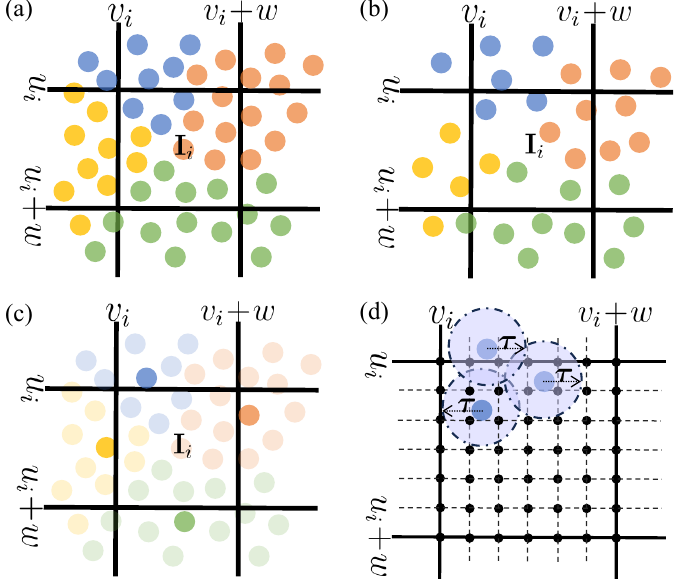} 
\caption{\textbf{Some toy examples for analysing challenges.} The solid black line represents the boundary of pixel patches and the colors represent different point sets. (a) and (b) are two cases with dense and sparse points respectively. The points from different sets are projected into the same pixel patch. (c) A case of GeoLoc \cite{Geoloc}. The brighter points represent the set centers. The pixel patch containing a center point is defined as the sole patch correlated with the corresponding set. (d) A case of 2D3D-MATR \cite{li20232d3d}. The pixels (black) with a distance of less than $\boldsymbol{\tau}$ from the projected points (blue) are overlapping pixels. If the point-pixel overlap ratio in a set and a patch is greater than $\boldsymbol{\tau}_{o}$, the correlation is determined to be 1, otherwise 0. It is usually biased by the resolution gap.}
\label{fig2}
\end{figure}

\subsection{Challenges}
Due to the significant modality gaps, at least three key aspects need to be considered.

\vspace{1mm}
\textbf{Misalignment.} Without any semantic priors, we generate pixel patches and point sets with distance-based clustering in 2D and 3D space respectively. As shown in Fig. \ref{fig2} (a), the projections of point sets and the pixel patches often overlap each other, so their correlations are not discrete binary values. 

\vspace{1mm}
\textbf{Point Cloud Density.} In fact, due to the diverse densities of a point cloud across different areas, the point sets always contain a different number of 3D points, as shown in Fig. \ref{fig2} (b). We argue that a good correlation should reflect this fact.


\vspace{1mm}
\textbf{Data Resolution Gap.} Point clouds and images describe the same object at different resolutions. For example, in Fig. \ref{fig1}, the point cloud records a tree with about 150 LiDAR points while the image records it with about 1600 pixels. We argue that the correlation of a point set with a pixel patch should exclude this negative effect.

In the concurrent works, both GeoLoc \cite{Geoloc} and 2D3D-MATR \cite{li20232d3d} adopt binary values to describe the correlation between a point set and a pixel patch, as illustrated in Fig. \ref{fig2} (c) and (d). Analogous to the concepts in knowledge distillation \cite{gou2021knowledge}, these correlations can be seen as hard targets, expressing limited information. Besides, they fail to address the aforementioned challenges simultaneously. 


\subsection{Our Quantity-Aware Strategy}
\label{sec:cardiaware}
Departing from the binary match-or-not approach, we interpret the matching of point sets and pixel patches as an optimal transport problem \cite{cuturi2013sinkhorn, detone2018superpoint, sarlin2020superglue}. Specifically, we model the correlation degree of each point set as an allocatable item, with pixel patches as distribution receptacles. This allows a point set to simultaneously exhibit positive correlation with several pixel patches, reflecting their inherent spatial misalignment. Besides, we employ continuous values ranging from 0 to 1 to represent the magnitude of correlation.

To address the data resolution gap and varying point cloud densities, we opt to quantify correlations based solely on the proportions of 3D points. This is achieved from a bilateral perspective. One side is point set and the other is pixel patch. As illustrated in Fig. \ref{fig2} (a), denote the vertical range of the pixel patch $\textbf{I}_i$ as $\boldsymbol{\Gamma}^{u}_i \! = \!\{u\,\big|\,u\in\mathbb{R}, u_i \leq u < u_i \! + \! w\}$ and the horizontal range as $\boldsymbol{\Gamma}^{v}_i \! = \! \{v\,\big|\,v\in\mathbb{R}, v_i \leq v < v_i \! + \! w\}$, where $w$ is the width and height of a pixel patch. We compute the proportion of 3D points in the point set $G^{\textbf{P}}_j$ that can be projected into the pixel patch $\textbf{I}_i$ as:
\begin{align}
    \label{eq3}
	r_{\overleftarrow{ij}} = \frac{\Big|\{ \tilde{\mathbf{p}} \in  G^{\textbf{P}}_j \,\big|\, [\tilde{u},\tilde{v}]^{\mathsf{T}} = \mathcal{F}_{p}(\textbf{K}\overline{\textbf{T}}\tilde{\mathbf{p}}), \tilde{u} \in \boldsymbol{\Gamma}^{u}_i, \tilde{v} \in \boldsymbol{\Gamma}^{v}_i\}\Big|}{|G^{\textbf{P}}_j|},
\end{align}
where $\overline{\textbf{T}}$ is the ground-truth camera pose and $|\cdot|$ denotes the set cardinality. On the contrary, we compute the proportion of 3D points in $\textbf{I}_i$ that are projected from the point set $G^{\textbf{P}}_j$ as:
\begin{align}
    \label{eq4}
	r_{\overrightarrow{ij}} = \frac{\Big|\{ \tilde{\mathbf{p}} \in  G^{\textbf{P}}_j \,\big|\, [\tilde{u},\tilde{v}]^{\mathsf{T}} = \mathcal{F}_{p}(\textbf{K}\overline{\textbf{T}}\tilde{\mathbf{p}}), \tilde{u} \in \boldsymbol{\Gamma}^{u}_i, \tilde{v} \in \boldsymbol{\Gamma}^{v}_i\}\Big|}{\Big|\{ \tilde{\mathbf{p}} \in  \textbf{P} \,\big|\, [\tilde{u},\tilde{v}]^{\mathsf{T}} = \mathcal{F}_{p}(\textbf{K}\overline{\textbf{T}}\tilde{\mathbf{p}}), \tilde{u} \in \boldsymbol{\Gamma}^{u}_i, \tilde{v} \in \boldsymbol{\Gamma}^{v}_i\}\Big|}.
\end{align}
Note that the calculations of $r_{\overleftarrow{ij}}$ and $r_{\overrightarrow{ij}}$ exclude the number of 2D pixels. All numerators and denominators are either the cardinality of a 3D point set or the cardinality of the 3D projected point set within the pixel patch. This is the origin of the term quantity-aware.

Based on the above bilateral proportions, we utilize an aggregation function $\boldsymbol{\sigma}(\cdot)$ to generate a continuous value as the degree of correlation between $\textbf{I}_i$ and $G^{\textbf{P}}_j$. Thus, we can get a correlation matrix with $N_I \! \times \! N_q$ elements between all point sets and pixel patches. Since the FoVs (i.e., field of view) of point clouds and images are not perfectly aligned, some point sets and pixel patches will not be correlated with each other. For the completeness of the optimal transport model \cite{cuturi2013sinkhorn}, we introduce two virtual receptacles to accommodate the pixel patches and point sets that exhibit no correlation. Finally, we calculate an augmented correlation matrix $\textbf{W}^{c} \in \mathbb{R}^{(N_I + 1)\times(N_q + 1)}$ as:
\begin{equation}
\label{eq5}
\textbf{W}^{c}_{ij} \! = \!
    \begin{cases}
        \boldsymbol{\sigma}(r_{\overleftarrow{ij}}, r_{\overrightarrow{ij}})\,, & i \! \leq \! N_I \land j \! \leq \! N_q,\\
	    1 - \sum_{j'}\,r_{\overrightarrow{ij'}}\,, & i \! \leq \! N_I \land j \! = \! N_q + 1,\\
        1 - \sum_{i'}\,r_{\overleftarrow{i'j}}\,, & i \! = \! N_I \! + \! 1 \land j \! \leq \! N_q,\\
        0\,, & \mathrm{otherwise},
    \end{cases}
\end{equation}
where the last row and column are the correlations towards the virtual receptacles. To find a more effective $\boldsymbol{\sigma}(\cdot)$, we will evaluate three forms of function including $\mathrm{min}(\cdot), \mathrm{max}(\cdot)$ and $\mathrm{mean}(\cdot)$ in ablation studies. It turns out that $\mathrm{min}(\cdot)$ is the best. We will display the results there.

\begin{figure*}[t]
\centering
\includegraphics[width=1\textwidth]{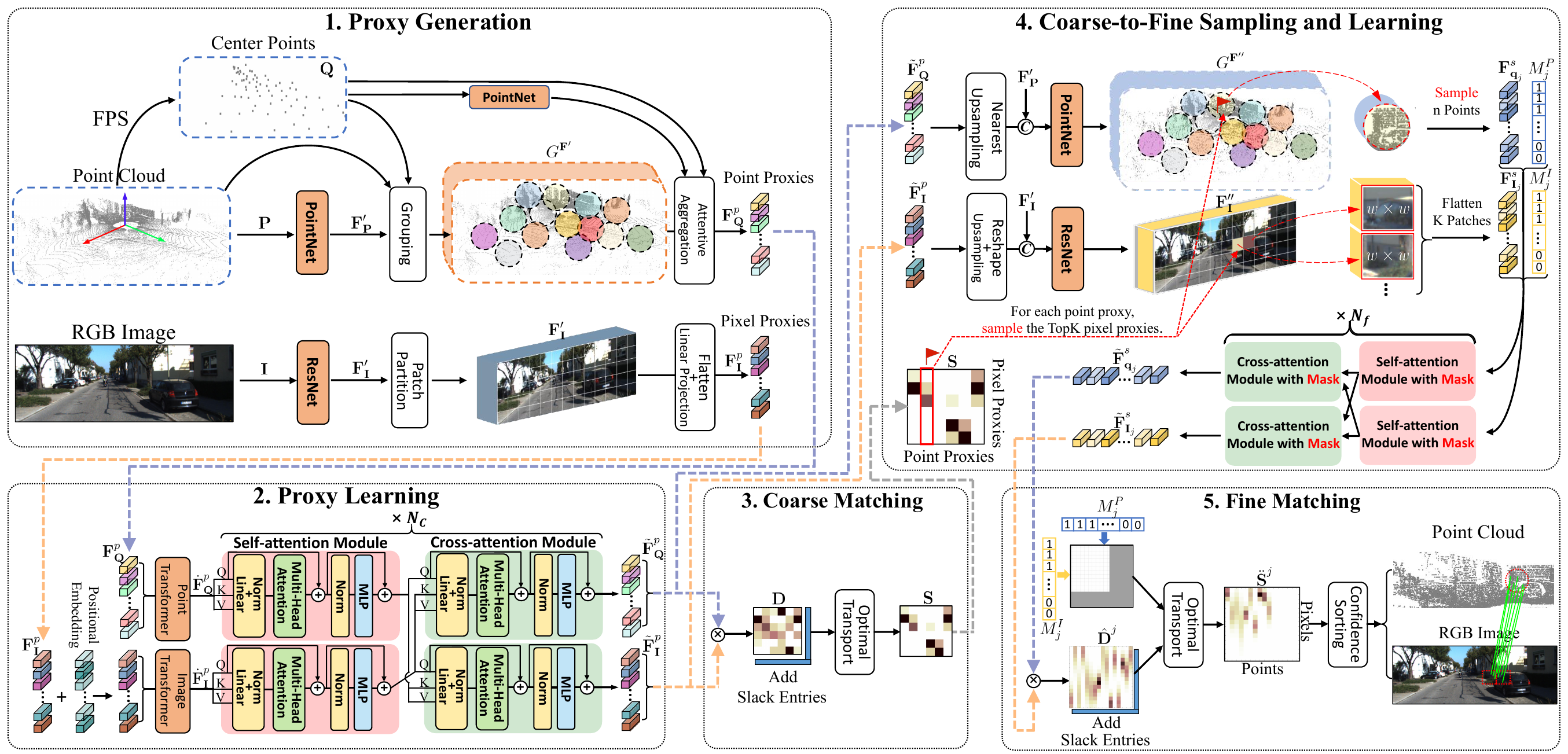}
\caption{\textbf{Overview of the proposed CFI2P Framework}. 1) The image and point cloud are divided into many non-overlapping regions to extract the local proxies of point sets and pixel patches. 2) Hybrid Transformers and cross-attention are adopted to capture global and cross-modal contexts between the proxies. 3) A differentiable optimal transport algorithm is adopted to match the proxies. 4) After fusing the rich contexts to point and pixel level, we sample n points within the set of each candidate point proxy, and select its top k pixel proxies. Binary sampling masks are produced to guide the subsequent learning. 5) The point-to-pixel correspondences are established by the fine level optimal transport algorithm and confidence sorting strategy.}
\label{fig3}
\end{figure*}

\section{Coarse-to-Fine Framework}
\label{sec:framework}
We design a coarse-to-fine framework, named CFI2P, to first establish the quantity-aware set-to-patch correspondences at the coarse level. After that, we refine the correspondences to the point and pixel levels according to the learned quantity-aware priors. The overall architecture is depicted in Fig. \ref{fig3}, and we will introduce the details as follows.

\subsection{Proxy Generation}
To make the 2D image $\textbf{I}$ and the 3D point cloud $\textbf{P}$ suitable for coarse-to-fine mechanism, the first step is to extract coarse representations, namely \textit{pixel proxy} and \textit{point proxy}. 

For images, we use a light-weight ResNet \cite{he2016deep} to extract feature $\textbf{F}_{\textbf{I}}' \in \mathbb{R}^{H' \times W' \times d_f}$. Since images have a compact and regular grid data structure, we follow the standard Vision Transformer \cite{dosovitskiyimage} to split $\textbf{I}$ into non-overlapping patches $\{\textbf{I}_i\}_{i=1}^{N_I}$ with a resolution of $w \! \times \! w$, and then use a linear projection layer to project their features as pixel proxies $\textbf{F}_{\textbf{I}}^{p} \in \mathbb{R}^{N_I \times d_p}$, where $N_I \! = \! H'W'/w^{2}$ and $d_p$ represents the feature dimension. 

Since point clouds are unordered and irregular, a light-weight PointNet \cite{qi2017pointnet, qi2017pointnet++} is adopted to extract point-wise features $\textbf{F}_{\textbf{P}}' \in \mathbb{R}^{N \times d_f}$. Simultaneously, we conduct Farthest Point Sampling (FPS) to obtain a fixed number of points $\textbf{Q} = \{\textbf{q}_1,\textbf{q}_2,\dots,\textbf{q}_{N_q}\}$, which will serve as the centers of some local regions. The associated features are denoted as $\textbf{F}_{\textbf{Q}}' \in \mathbb{R}^{N_q \times d_f}$. Then, we take a point-to-node grouping \cite{li2018so} strategy to assign every point in $\textbf{P}$ to its nearest center in $\textbf{Q}$. After grouping, the raw point cloud $\textbf{P}$ is split into $N_q$ point sets $\{G^{\textbf{P}}_{j}\}_{j=1}^{N_q}$ and the point-wise feature
$\textbf{F}_{\textbf{P}}'$ is also split into $N_q$ feature sets $\{G^{\textbf{\textbf{F}'}}_{j}\}_{j=1}^{N_q}$ as:
\begin{align}
\label{eq6}
\left\{
	\begin{aligned}
	G^{\textbf{P}^{\,}}_{j} &= \, \{\textbf{p}_i \in \textbf{P} \,\big| \, \Vert \textbf{p}_i - \textbf{q}_j \Vert \leq \Vert \textbf{p}_i - \textbf{q}_{k} \Vert, k \neq j \}\\
	G^{\textbf{F}'}_{j} &= \, \{ \textbf{F}'_{\textbf{p}_i} \in \textbf{F}'_\textbf{P} \,\big| \, \textbf{p}_i \in G^{\textbf{P}}_{j} \}
	\end{aligned}
\right. .
\end{align}
Note that different $G^{\textbf{P}^{\,}}_{j}$ holds different numbers of points. To enhance the local geometry representation, we modified a standard Point Transformer \cite{zhao2021point} layer with scattered indexing to aggregate unified point proxies for these heterogeneous sets $G^{\textbf{P}}$ and $G^{\textbf{F}'}$, named Attentive Aggregation. Specifically, $\textbf{q}_j$ and its features $\textbf{F}'_{\textbf{q}_j}$ are linearly projected as the \textit{Query} vectors, while the grouped points $G^{\textbf{P}}_j$ and features $G^{\textbf{F}'}_{j}$ are linearly projected as the \textit{Key} and \textit{Value} vectors. We get the point proxies $\textbf{F}_{\textbf{Q}}^{p} \in \mathbb{R}^{N_q \times d_p}$ where a single proxy $\textbf{F}_{\textbf{q}_j}^{p} \in \mathbb{R}^{d_p}$ is computed with vector attention \cite{zhao2021point} as:
\begin{align}
    \label{eq7}
	\textbf{F}_{\textbf{q}_j}^{p} \! = \! \sum_{\textbf{p}_i \in G^{\textbf{P}}_j} \! \rho\Big(\gamma(\varphi(\textbf{F}'_{\textbf{q}_j}) - \psi(\textbf{F}'_{\textbf{p}_i})+\delta)\Big) \! \odot \! \Big(\alpha(\textbf{F}'_{\textbf{p}_i})+\delta\Big),
\end{align}
where $\delta \in \mathbb{R}^{d_p}$ is point position embedding $\theta(\textbf{q}_j \! - \! \textbf{p}_i)$, projecting the relative position to feature space. $\varphi, \psi$ and $\alpha$ are linear projection functions to generate \textit{Query}, \textit{Key} and \textit{Value}. $\gamma$ produces an attention weight vector for each \textit{Query}-\textit{Key} pair to modulate individual feature channels. $\rho$ is a \textit{softmax} function that operates on each point set $G^{\textbf{P}}_j$ to normalize the attention weights. All the functions keep the number of feature channels as ${d_p}$. $\odot$ represents Hadamard product.

\subsection{Proxy Learning}
So far, the proxies $\textbf{F}_{\textbf{I}}^{p}$ and $\textbf{F}_{\textbf{Q}}^{p}$ only encode their local geometry, but they both lack global and cross-modal contexts.

Following the standard Vision Transformer, we treat $\textbf{F}_{\textbf{I}}^{p}$ as a 1D sequence and add sinusoidal position embedding \cite{vaswani2017attention}. Then, standard scalar dot-product attention \cite{dosovitskiyimage} is utilized to enhance $\textbf{F}_{\textbf{I}}^{p}$ with global contexts as:
\begin{align}
    \label{eq8}
	\dot{\textbf{F}}_{\textbf{I}}^{p} = \textbf{F}_{\textbf{I}}^{p} + \rho\big(\,\varphi(\textbf{F}_{\textbf{I}}^{p})\psi(\textbf{F}_{\textbf{I}}^{p})^{\mathsf{T}}\,\big) \alpha\big(\textbf{F}_{\textbf{I}}^{p}\big),
\end{align}
where $\varphi, \psi$ and $\alpha$ generate \textit{Query}, \textit{Key} and \textit{Value} as \begin{footnotesize}$\varphi(\textbf{F}_{\textbf{I}}^{p})$, $\psi(\textbf{F}_{\textbf{I}}^{p})$, $\alpha(\textbf{F}_{\textbf{I}}^{p}) \in \mathbb{R}^{N_I \times d_p}$ \end{footnotesize}. $\rho$ is a \textit{softmax} function to normalize the attention weight scalars of each \textit{Query}-\textit{Key} pair. We capture global contexts for the point proxies $\textbf{F}_{\textbf{Q}}^{p}$ in the same way. The global-aware point proxies are denoted as $\dot{\textbf{F}}^{p}_{\textbf{Q}}$.

To exchange information between the enhanced $\dot{\textbf{F}}_{\textbf{I}}^{p}$ and $\dot{\textbf{F}}_{\textbf{Q}}^{p}$, we alternately perform self-attention and cross-attention. The acquired proxies with cross-modal contexts are denoted as $\tilde{\textbf{F}}^{p}_{\textbf{Q}} \in \mathbb{R}^{N_q \times d_p}$ and $\tilde{\textbf{F}}_{\textbf{I}}^{p} \in \mathbb{R}^{N_I \times d_p}$.

\subsection{Coarse Matching}
\label{sec:cmatch}
By matching pixel and point proxies, this module proposes the coarse correspondences between pixel patches and point sets. As mentioned in Section \ref{sec:cardiaware}, the matching problem is interpreted as an optimal transport problem \cite{sarlin2020superglue}. So, we first compute a pairwise distance matrix $\textbf{D} \in \mathbb{R}^{N_I \times N_q}$ as:
\begin{align}
    \label{eq9}
	\textbf{D} = \tilde{\textbf{F}}_{\textbf{I}}^{p}(\tilde{\textbf{F}}^{p}_{\textbf{Q}})^{\mathsf{T}} / \sqrt{d_p}\,,
\end{align}
where each entry $\textbf{D}_{ij}$ measures the cost of assigning a pixel proxy $\tilde{\textbf{F}}_{\textbf{I}_i}^{p}$ to a point proxy $\tilde{\textbf{F}}^{p}_{\textbf{q}_j}$. Synchronize with Eq. \ref{eq5}, we augment the matrix $\textbf{D}$ with an additional row and column of slack entries, so that the point sets (pixel patches) that cannot be assigned to any pixel patches (point sets) can be assigned to the slack entries. On the augmented matrix, we run the Sinkhorn algorithm \cite{sinkhorn1967concerning} to compute a soft assignment matrix $\tilde{\textbf{S}} \in \mathbb{R}^{(N_I+1) \times (N_q+1)}$. After convergence, the slack entries are removed from $\tilde{\textbf{S}}$ to get a matching score matrix $\textbf{S} \in \mathbb{R}^{N_I \times N_q}$.

\subsection{Coarse-to-Fine Sampling and Learning}
To fuse global and cross-modal contexts to a finer resolution, we concatenate the upsampled proxies and fine-level features, and then fuse them with a ResNet and PointNet to get $\textbf{F}_{\textbf{I}}'' \in \mathbb{R}^{H' \times W' \times d_f}$ and $\textbf{F}_{\textbf{P}}'' \in \mathbb{R}^{N \times d_f}$. Note that each point proxy $\tilde{\textbf{F}}^{p}_{\textbf{q}_j}$ is only replicated $|G^{\textbf{P}}_{j}|$ times and fused with the features $G^{\textbf{F}'}_{j}$. $\textbf{F}_{\textbf{P}}''$ is also grouped as $G^{\textbf{F}''}$ according to Eq. \ref{eq6}.

According to the coarse score matrix $\textbf{S}$, the $j$-th point set that satisfies $\sum_{i=1}^{N_I} \textbf{S}_{ij} \! > \! 0$ will recognized as the candidates for further refinement. To facilitate batch processing, we first resample $n$ points in each candidate set $G^{\textbf{P}}_{j}$ and the associated features are denoted as $\textbf{F}^{s}_{\textbf{q}_j} \in \mathbb{R}^{n \times d_f}$. Another byproduct of the sampling prior that is recorded as a binary point mask $M^{P}_j \in \mathbb{R}^{n \times 1}$, where the first $|G^{\textbf{P}}_{j}|$ items are 1 and the last $n \! - \! |G^{\textbf{P}}_{j}|$ items are 0 if $|G^{\textbf{P}}_{j}| \! < \! n$. Besides, we select the pixel patches (with a resolution of $w \times w$) with the top $k$ highest items in the $j$-th column of $\textbf{S}$. The associated features are $\textbf{F}^{s}_{\textbf{I}_{j}} \in \mathbb{R}^{m \times d_f}$ where $ m \! = \! k \! \times \! w^2$. Meanwhile, it also produces a similar binary pixel mask $M^{I}_j \in \mathbb{R}^{m \times 1}$ where the first $k' \! \times \! w^2$ items are 1 and the last $(k \! - \! k') \! \times \! w^2$ items are 0 if there are only $k'$ items $>0$ in the $j$-th column of $\textbf{S}$ and $k' < k$.

Afterwards, we will extract interactive features between $\textbf{F}^{s}_{\textbf{q}_j}$ and $\textbf{F}^{s}_{\textbf{I}_{j}}$. To eliminate the negative effect of invalid sampled points and pixels (i.e., the points and pixels filling data batches), we embed the above masks into the attention mechanism for focusing. Take the cross-attention module with masks as an example, we first replicate $M^{P}_j$ and $M^{I}_j$ as matrices $\textbf{M}^{\textbf{P}}_j \in \mathbb{R}^{n \times d_f}$ and $\textbf{M}^{\textbf{I}}_j \in \mathbb{R}^{m \times d_f}$. The attention weight $\textbf{W}_a \in \mathbb{R}^{m \times n}$ between $\textbf{F}^{s}_{\textbf{q}_j}$ and $\textbf{F}^{s}_{\textbf{I}_{j}}$ is calculated as:
\begin{align}
\label{eq10}
    \textbf{W}_a = \rho\Big(\big(\textbf{M}^{\textbf{I}}_j \odot \varphi(\textbf{F}^{s}_{\textbf{I}_{j}})\big) \big(\textbf{M}^{\textbf{P}}_j \odot \psi(\textbf{F}^{s}_{\textbf{q}_j})\big)^{\mathsf{T}}\Big),
\end{align}
and we fuse the information from $\textbf{F}^{s}_{\textbf{q}_j}$ to $\textbf{F}^{s}_{\textbf{I}_{j}}$ as:
\begin{align}
\label{eq11}
    \tilde{\textbf{F}}^{s}_{\textbf{I}_{j}} = \textbf{F}^{s}_{\textbf{I}_{j}} + \textbf{W}_a\left(\textbf{M}^{\textbf{P}}_j \odot \alpha(\textbf{F}^{s}_{\textbf{q}_j})\right),
\end{align}
where $\varphi, \psi$ and $\alpha$ follow the definitions in Eq. \ref{eq8} to generate \textit{Query, Key} and \textit{Value}. $\rho$ is a \textit{softmax} function. The acquired point-wise and pixel-wise features with hybrid information are denoted as $\tilde{\textbf{F}}^{s}_{\textbf{q}_j} \in \mathbb{R}^{n \times d_f}$ and $\tilde{\textbf{F}}^{s}_{\textbf{I}_{j}} \in \mathbb{R}^{m \times d_f}$.

\subsection{Fine Matching}
\label{sec:fine}
This module first performs point-to-pixel matching using the resampling priors, followed by a confidence sorting strategy with quantity-aware priors to select better correspondences.

\vspace{1mm}
\textbf{Matching.} We integrate the sampling masks $M^{P}_j$ and $M^{I}_j$ into the pairwise distance matrix $\hat{\textbf{D}}^j \in \mathbb{R}^{m \times n}$ as:
\begin{align}
    \label{eq12}
	\hat{\textbf{D}}^j = \Big(M^{I}_j(M^{P}_j)^{\mathsf{T}}\Big) \odot \left(\tilde{\textbf{F}}^{s}_{\textbf{I}_{j}}(\tilde{\textbf{F}}^{s}_{\textbf{q}_j})^{\mathsf{T}}\right) / \sqrt{d_f}.
\end{align}
Since some points will not be associated with any pixels and vice versa, we also augment matrix $\hat{\textbf{D}}^j$ with an additional row and column of slack entries, and run the Sinkhorn algorithm to compute the soft assignment matrix $\hat{\textbf{S}}^j \in \mathbb{R}^{(m+1) \times (n+1)}$. We remove the slack entries to get a point-to-pixel matching score matrix $\ddot{\textbf{S}}^j \in \mathbb{R}^{m \times n}$. 

\begin{figure}[t]
\centering
\includegraphics[width=0.45\textwidth]{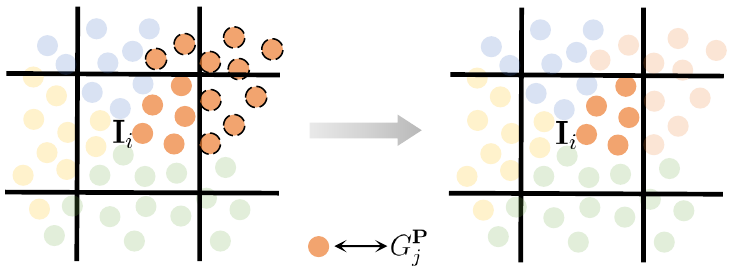} 
\caption{\textbf{Target of our confidence sorting strategy}. Assuming that the point set $G^{\textbf{P}}_j$ and the pixel patch $\textbf{I}_i$ are coarsely matched, we tend to select the points in $G^{\textbf{P}}_j$ that can be projected into $\textbf{I}_i$.}
\label{fig4}
\end{figure}

\vspace{1mm}
\textbf{Sorting.} Without loss of generality, suppose we only select the top 1 highest item (e.g., the $i$-th item) in the $j$-th column of the coarse score matrix $\textbf{S}$. It means that the point set $G^{\textbf{P}}_j$ is matched with the pixel patch $\textbf{I}_i$. Aiming at mitigating the misalignment problem as illustrated in Fig. \ref{fig4}, we first sort the sum of each column in the fine score matrix $\ddot{\textbf{S}}^j$ as:
\begin{align}
    \label{eq13}
	\mathrm{sort}(\sum\nolimits_{i'}\ddot{\textbf{S}}^j_{i'1}\,,\, \sum\nolimits_{i'}\ddot{\textbf{S}}^j_{i'2}\,,\, \dots\,,\, \sum\nolimits_{i'}\ddot{\textbf{S}}^j_{i'n}), 
\end{align}
where each item $\sum\nolimits_{i'}\ddot{\textbf{S}}^j_{i'j'}$ represents the confidence score that the $j'$-th point can be projected into $\textbf{I}_i$. Note that the coarse matching score $\textbf{S}_{ij}$ quantifies the correlation between $\textbf{I}_i$ and $G^{\textbf{P}}_j$. According to the quantity-aware priors in Sections \ref{sec:cardiaware}, the predicted score $\textbf{S}_{ij}$ is proportional to the ratio of points in $G^{\textbf{P}}_j$ that can be projected into $\textbf{I}_i$. Thus, we calculate the expectation of the point number as:
\begin{align}
    \label{eq14}
    N^j = |G^{\textbf{P}}_j| \cdot \textbf{S}_{ij}\,.
\end{align}
Afterwards, we preserve the $N^j$ points with higher confidence. 

Thus, we will get $N^j$ point-to-pixel correspondences for the point set $G^{\textbf{P}}_j$ via matching and sorting. We repeat the two operations on the resampled points and pixels of all candidate point sets. Finally, all the point-to-pixel correspondences are collected for image-to-point cloud registration.

\subsection{Supervision}
The coarse and fine parts of CFI2P are trained separately.

\vspace{1mm}
\textbf{Coarse-level Supervision.} Following Eq. \ref{eq5}, we quantify the correlations between the $N_I$ pixel patches and the $N_q$ point sets. The acquired correlation matrix with slack entries is denoted as $\textbf{W}^{c} \in \mathbb{R}^{(N_I + 1)\times(N_q + 1)}$. We aim for the learned geometric features at the set and patch levels, namely point and pixel proxies, to implicitly reflect these correlations. For this purpose, we supervise the soft assignment matrix $\tilde{\textbf{S}}$ with a weighted negative log-likelihood loss \cite{sarlin2020superglue, yu2021cofinet} as:
\begin{equation}
    \label{eq15}
	\mathcal{L}_c = \frac{-\sum_{i,j} \textbf{W}^{c}_{ij} \log(\tilde{\textbf{S}}_{ij})}{\sum_{i,j} \textbf{W}^{c}_{ij}}.
\end{equation}

\vspace{1mm}
\textbf{Fine-level Supervision.} We design a projected point distance loss with the sampling masks to supervise the the soft assignment matrix $\hat{\textbf{S}}^{\ell}$ of the $\ell$-th candidate point set. Denote the resampled 3D points as $\{\textbf{p}^{\ell}_{1}, \dots, \textbf{p}^{\ell}_{n}\}$, the point mask as $M^{P}_\ell \in \mathbb{R}^{n \times 1}$, the 2D coordinates of the resampled pixels as $\{\textbf{u}^{\ell}_{1}, \dots,\textbf{u}^{\ell}_{m}\}$, and the pixel mask as $M^{I}_\ell \in \mathbb{R}^{m \times 1}$. We define a binary matrix $\textbf{W}^{\ell} \in \mathbb{R}^{(m + 1)\times(n + 1)}$ where any $\textbf{W}^{\ell}_{ij}$ subject to $i \leq m$ and $j \leq n$ is computed as:
\begin{equation}
\label{eq16}
\textbf{W}^{\ell}_{ij} = 
        \big\llbracket \Vert \textbf{u}^{\ell}_{i} - \mathcal{F}_{p}(\textbf{K}\overline{\textbf{T}}\textbf{p}^{\ell}_{j})\Vert \leq \boldsymbol{\tau} \big\rrbracket,
\end{equation}
where $\llbracket\cdot\rrbracket$ is the Iversion bracket, and $\boldsymbol{\tau}$ is a distance threshold to classify the positive pixels relative to a projected point, as illustrated in Fig. \ref{fig2} (d). Synchronized with Eq. \ref{eq12}, we then mask the above $m\!\times\!n$ elements in $\textbf{W}^{\ell}$ with $M^{I}_\ell(M^{P}_\ell)^{\mathsf{T}}$. Other elements on the edges in $\textbf{W}^{\ell}$ are then computed as:
\begin{equation}
\label{eq17}
\textbf{W}^{\ell}_{ij}\!=\! 
    \begin{cases}
	\big\llbracket \sum_{i'=1}^{m} \textbf{W}^{\ell}_{i'j} < 1 \big\rrbracket, \!& i\!=\!m+1 \land j\!\leq\!n,\\
        \big\llbracket \sum_{j'=1}^{n} \textbf{W}^{\ell}_{ij'} < 1 \big\rrbracket, \!& i\!\leq\!m \land j\!=\!n\!+\!1,\\
        0, & i\!=\!m\!+\!1 \land j\!=\!n\!+\!1.
    \end{cases}
\end{equation}
Finally, the fine-level loss is the sum of the weighted negative log-likelihood loss across all candidate point sets as:
\begin{equation}
    \label{eq18}
	\mathcal{L}_f = \sum_{\ell} \frac{-\sum_{i,j} \textbf{W}^{\ell}_{ij}\log(\hat{\textbf{S}}^{\ell}_{ij})}{\sum_{i,j} \textbf{W}^{\ell}_{ij}}.
\end{equation}

\begin{table*}[t]
\centering
\caption{Quantitative results of point-to-pixel correspondences. The best results are highlighted in bold. CFI2P+X denotes the hybrid method that takes CFI2P as the backbone and utilizes the supervision of X for set-to-patch matching at the coarse level. Here, we report the IR(\%)/FMR(\%) with different thresholds $\tau_d$(pixel)/$\tau_m$(\%) .}
\renewcommand\arraystretch{1.2}
\begin{tabular}{c|ccc|ccc}
\hline \hline
 & \multicolumn{3}{c|}{KITTI Odometry}                                         & \multicolumn{3}{c}{NuScenes}                                               \\ \hline
 & $\tau_d$=1\,/\,$\tau_m$=0.2 & $\tau_d$=2\,/\,$\tau_m$=0.2 & $\tau_d$=3\,/\,$\tau_m$=0.2 & $\tau_d$=1\,/\,$\tau_m$=0.2 & $\tau_d$=2\,/\,$\tau_m$=0.2 & $\tau_d$=3\,/\,$\tau_m$=0.2 \\ \hline
DeepI2P-GridCls (CVPR'21) \cite{li2021deepi2p} & 02.29 / 00.00        & 09.25 / 00.00        & 20.21 / 58.61           & 02.73 / 00.00        &  11.08 / 00.09    & 24.99 / 80.17          \\
CorrI2P (TCSVT'23) \cite{ren2022corri2p} & 10.84 / 18.12       & 27.69 / 64.60          &  42.18 / 82.38         & 17.82 / 41.41         & 37.48 / 85.78        & 51.99 / 93.47          \\ \hline
CFI2P + GeoLoc (RA-L'23) \cite{Geoloc} & 14.41 / 20.86         & 34.19 / 89.65        &   49.40 / 95.99        & 29.24 / 79.88       & 53.08 / 93.90        &  66.96 / 97.05         \\
CFI2P + 2D3D-MATR (ICCV'23) \cite{li20232d3d} &  17.14 / 42.32        &  41.07 / 93.36    & 58.90 / 97.19       & 29.16 / 79.35      & 52.98 / 94.02   &  66.89 / 97.05         \\
 CFI2P + Quantity-Aware (ours) & \textbf{23.25} / \textbf{67.14}        & \textbf{51.10} / \textbf{94.47}        & \textbf{69.16} / \textbf{97.37}          & \textbf{30.58} / \textbf{82.62}         & \textbf{57.65} / \textbf{95.25}       & \textbf{74.16} / \textbf{97.71}          \\ \hline \hline
\end{tabular}     
\label{tab1}
\end{table*}

\section{Experiments}
\label{sec:experiment}
\subsection{Dataset} We have conducted experiments on two large-scale outdoor benchmarks, i.e., the KITTI Odometry \cite{geiger2013vision} and NuScenes \cite{caesar2020nuscenes} datasets. To construct the image-to-point cloud registration problem, it should be ensured that the image captures a part of the same scene in the point cloud. Thus, we follow the same data protocol in the previous works \cite{li2021deepi2p, ren2022corri2p} to get image-point cloud pairs for a fair comparison:

\vspace{1mm}
\noindent\textbf{KITTI Odometry.} There are 11 data sequences that include the calibration parameters in this dataset. The data sequences 0-8 are used for training while 9-10 for testing. We first select the image and point cloud with same frame ID. Then we sample 40960 points per point cloud and resize the image as $160\!\times\!512\!\times3$. After that, we generate a random transformation to rotate and translate the point cloud. It involves a rotation around the up-axis within a range of 360°, and a translation on the ground within a range of 10 meters. For image-to-point cloud registration, the ground-truth camera poses are calculated with the inverse transformation.

\vspace{1mm}
\noindent\textbf{NuScenes.} We use the official SDK to split 850 scenes for training and 150 scenes for testing. Different from KITTI Odometry, we accumulate multiple frames of point cloud around the frame ID of an RGB image, which results in a more comprehensive 3D map of the scene. The point cloud is downsampled to 40960 points and the image is resized as $160\!\times\!320\!\times3$. Afterwards, we follow the same transformation method in KITTI Odometry to rotate and translate the point cloud. The ranges of rotation and translation are the same.

\begin{figure}[t]
\centering
\includegraphics[width=0.47\textwidth]{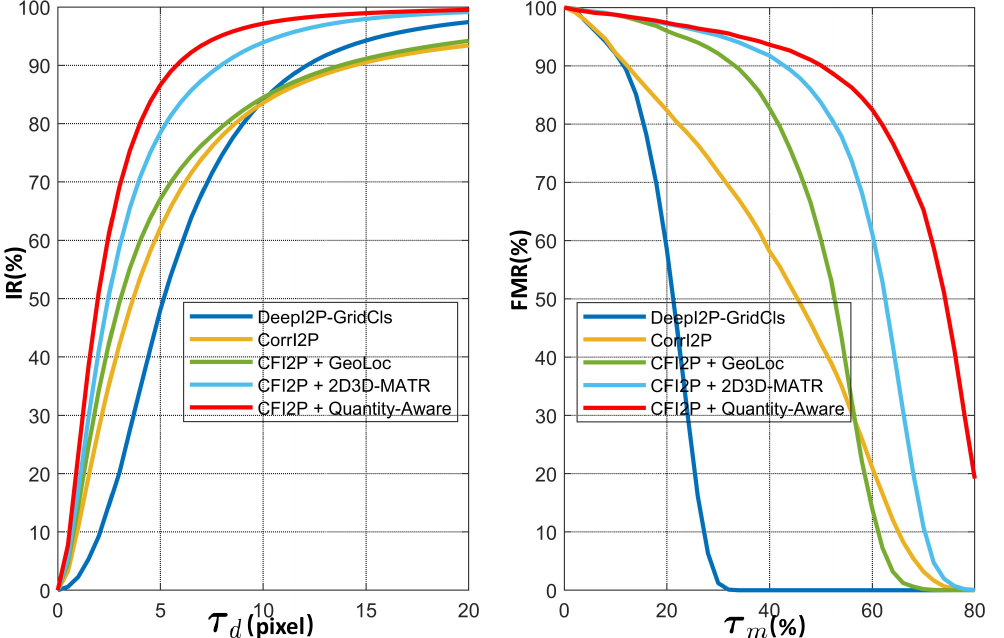} 
\caption{\textbf{Left}: The curves of IR across different thresholds. \textbf{Right}: The curves of FMR ($\tau_d$ for IR is 3 pixels) across different thresholds. We obtained the quantitative results on the KITTI Odometry dataset.}
\label{fig5}
\end{figure}  

\subsection{Implementation Details} 
We adopt the PyTorch 1.12 to implement the architecture. In the proxy generation, an image is split into patches with a resolution of $8\!\times\!8$ while a point cloud is split into 256 sets. So the number of pixel and point proxies are $(H' \! \times \! W') / 64$ and 256. In default settings, we resample $n\!=\!65$ points in each point set $G^{\textbf{P}}_j$ and select the top 3 pixel proxies (i.e., $m\!=\!3\!\times\!64$ pixels) in the coarse-to-fine sampling module. The distance threshold $\boldsymbol{\tau}$ in Eq. \ref{eq15} is 1 pixel. Additionally, considering that the pixel resolution and density of an image are much higher than the density of a point cloud, we will downsample the image as $1/4$ of the original dimension during forward propagation and then perform registration at the downsampled dimension. Thus, $H' \! \times \! W'$ are $40\times128$ and $40\times80$ in KITTI Odometry and NuScenes, respectively. We collect the point-to-pixel correspondences output by CFI2P and then use a RANSAC-based EPnP solver to estimate the camera poses. The number of RANSAC iterations is 500 and the threshold for inlier reprojection error is 1. 

We conduct experiments on an NVIDIA RTX 3090 GPU with an ADAM optimizer. In KITTI Odometry, we first train the coarse part of CFI2P for 60 epochs, with a batch size of 8. The initial learning rate is 1e-3 and multiplied by 0.8 every 6 epochs. Since the set-to-patch correspondences provide a good initialization for fine matching, we can train the fine part of CFI2P for only 2 epochs to achieve a fast convergence. In NuScenes, we first train the coarse part of CFI2P for 30 epochs, with a batch size of 8. The initial learning rate is 1e-3 and multiplied by 0.8 every 3 epochs. Additionally, the references for coarse-to-fine sampling are different during training and testing. We will sample pixels and points according to the quantified correlation matrix $\textbf{W}^{c}$ for training. By this way, the prediction error of coarse matching will not affect the training of fine matching. After convergence, we refer to the predicted coarse matching score matrix $\textbf{S}$ to perform coarse-to-fine sampling for testing.

\begin{table*}[t]
\centering
\caption{Quantitative results of image-to-point cloud registration. The best results are highlighted in bold. CFI2P+X denotes the hybrid method that takes CFI2P as the backbone and utilizes the supervision of X for set-to-patch matching at the coarse level. The mark $\star$ represents that the corresponding method is matching-free. Here, $\tau_r$ and $\tau_t$ for RR are 10° and 5 m, respectively. We report the RTE and RRE averaged on the successful registrations.}
\renewcommand\arraystretch{1.2}
\begin{tabular}{c|ccc|ccc}
\hline
\hline
\multicolumn{1}{c}{} & \multicolumn{3}{|c}{KITTI Odometry Dataset}                    & \multicolumn{3}{|c}{NuScenes Dataset}                            \\ \hline
                     & RR(\%)  & RTE(m)  & RRE(°)  & RR(\%)   & RTE(m)   & RRE(°)   \\ \hline
DeepI2P-GridCls (CVPR'21) \cite{li2021deepi2p}            & 80.18   & 1.137$\pm$0.696    & 5.883$\pm$2.013    & 62.67    & 2.224$\pm$1.094     & 7.370$\pm$1.647       \\
DeepI2P-3D (CVPR'21) \cite{li2021deepi2p} $\star$           & 38.34   & 1.206$\pm$0.748    & 6.083$\pm$2.294    & 18.82    & 1.730$\pm$0.969     & 7.058$\pm$2.184       \\
DeepI2P-2D (CVPR'21) \cite{li2021deepi2p} $\star$           & 74.50   & 1.417$\pm$0.897    & 3.877$\pm$2.687    & 92.53    & 1.949$\pm$1.050     & 3.017$\pm$2.285       \\
EFGHNet (RA-L'22) \cite{efghnet} $\star$           & 23.92   & 3.191$\pm$1.132    & 4.953$\pm$2.501    & 31.25    & 3.918$\pm$1.493     & 5.742$\pm$3.439       \\
CorrI2P (TCSVT'23) \cite{ren2022corri2p}              & 92.19   & 0.902$\pm$0.720    & 2.720$\pm$2.170    & 93.87    & 1.697$\pm$1.030     & 2.318$\pm$1.868        \\ \hline
CFI2P + GeoLoc (RA-L'23) \cite{Geoloc}              & 99.28   & 0.906$\pm$0.693    & 1.775$\pm$1.346    & 98.95    & 1.124$\pm$0.696     & 1.466$\pm$1.237        \\
CFI2P + 2D3D-MATR (ICCV'23) \cite{li20232d3d}              & 99.42   & 0.690$\pm$0.529   & 1.842$\pm$1.239    & 99.06    & 1.122$\pm$0.690    & \textbf{1.465}$\pm$1.230  \\
CFI2P + Quantity-Aware (ours)      & \textbf{99.44}   & \textbf{0.541$\pm$0.437}    & \textbf{1.383$\pm$1.085}    & \textbf{99.23}    & \textbf{1.093$\pm$0.668}     & 1.468$\pm$\textbf{1.211}   \\ \hline \hline       
\end{tabular}         
\label{tab2}
\end{table*}

\subsection{Evaluation Metrics} Similar to point cloud registration \cite{qin2022geometric}, we evaluate the matching accuracy of point-to-pixel correspondences with:

\vspace{1mm}
\noindent\textbf{Inlier Ratio (IR).} The fraction of inliers among all putative point-to-pixel correspondences. A correspondence is considered as an inlier if the distance between the projected point and the ground-truth pixel is smaller than a threshold $\boldsymbol{\tau}_d$. Denote the estimated correspondence set as $\tilde{C}$, IR is formulated as:
\begin{align}
    \label{eqA1}
	\mathrm{IR} &=  \frac{1}{|\tilde{C}|}\sum\nolimits_{(\textbf{u}_{i},\textbf{p}_{i}) \in \tilde{C}} \left\llbracket \Vert \textbf{u}_{i} - \mathcal{F}_{p}(\textbf{K}\overline{\textbf{T}}\textbf{p}_{i})\Vert <  \boldsymbol{\tau}_d \right\rrbracket,
\end{align}
where $\llbracket\cdot\rrbracket$ is the Iversion bracket. 


\vspace{1mm}
\noindent\textbf{Feature Matching Recall (FMR).} The fraction of data pairs whose inlier ratio is above a threshold $\boldsymbol{\tau}_m$. FMR measures the potential success rate during image-to-point cloud registration, which is formulated as:
\begin{align}
    \label{eqA2}
	\mathrm{FMR} &=  \frac{1}{M}\sum_{i=1}^{M} \left\llbracket \mathrm{IR}_i > \boldsymbol{\tau}_m \right\rrbracket,
\end{align}
where $M$ is the number of data pairs in the test dataset and $\mathrm{IR}_i$ is the \textit{Inlier Ratio} of the $i$-th data pair. 


Following DeepI2P \cite{li2021deepi2p}, we evaluate the image-to-point cloud registration performance with three metrics:

\vspace{1mm}
\noindent\textbf{Relative Rotation Error (RRE).} The mean of relative rotation error between the predicted (i.e., $\textbf{T} = [\textbf{R}|\textbf{t}] $) and ground-truth camera pose (i.e., $\overline{\textbf{T}} = [\bar{\textbf{R}}|\bar{\textbf{t}}] $). We first calculate the relative rotation matrix $\Ddot{\textbf{R}}$ between $\textbf{R}$ and $\bar{\textbf{R}}$ as:
\begin{align}
    \label{eqA3}
	\Ddot{\textbf{R}} &= \textbf{R}^{-1} \bar{\textbf{R}}.
\end{align}
Then, we convert $\Ddot{\textbf{R}}$ into three Euler angles (i.e., with respect to the X, Y and Z axes), and RRE is the sum of three angles.

\vspace{1mm}
\noindent\textbf{Relative Translation Error (RTE).} The mean of relative error in translation vectors between the predicted and the ground-truth camera pose as:
\begin{align}
    \label{eqA4}
	\mathrm{RTE} &= \Vert \textbf{t} - \bar{\textbf{t}} \Vert.
\end{align}

\vspace{1mm}
\noindent\textbf{Registration Recall (RR).} The fraction of successful registrations in the test dataset. A registration is considered as successful when the RRE is smaller than $\boldsymbol{\tau}_r$ and the RTE is smaller than $\boldsymbol{\tau}_t$. So, RR is formulated as:
\begin{align}
    \label{eqA5}
	\mathrm{RR} &=  \frac{1}{M}\sum_{i=1}^{M} \left\llbracket \mathrm{RRE}_i < \boldsymbol{\tau}_r \land \mathrm{RTE}_i < \boldsymbol{\tau}_{t} \right\rrbracket,
\end{align}
where $M$ is total number of the data samples. $\mathrm{RRE}_i$ and $\mathrm{RTE}_i$ are the relative rotation and translation error of the $i$-th data sample, respectively.

\begin{figure}[t]
\centering
\includegraphics[width=0.47\textwidth]{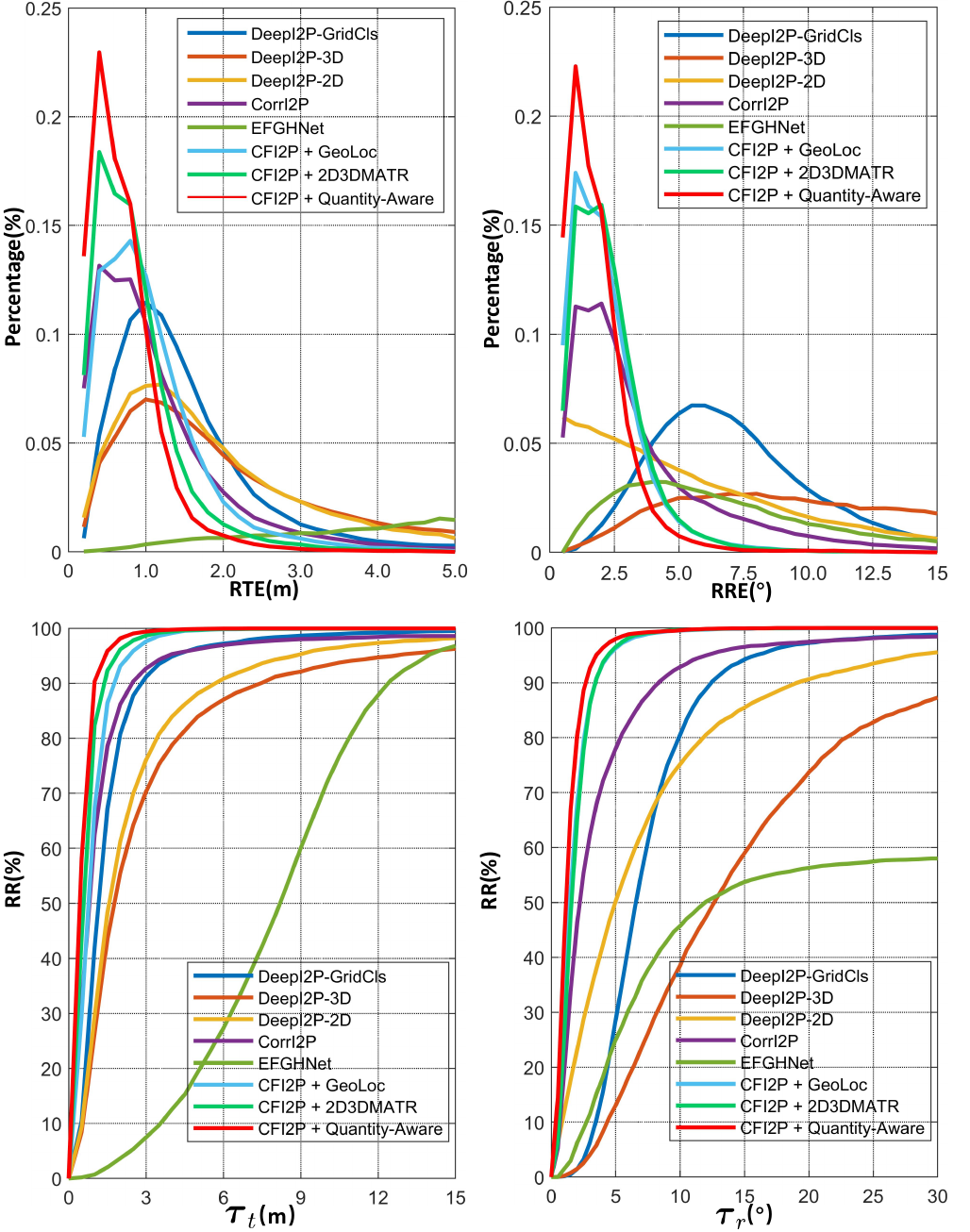} 
\caption{\textbf{Top Left}: The sample distribution of RTE results. \textbf{Top Right}: The sample distribution of RRE results. \textbf{Bottom Left}: The curves of RR across different RTE thresholds. \textbf{Bottom Right}: The curves of RR across different RRE thresholds. We obtained the results on the KITTI Odometry dataset.}
\label{fig6}
\end{figure}  

\subsection{Quantitative Results}
\textbf{Point-to-Pixel Correspondence.} We compare our method with the state-of-the-art matching-based methods, including DeepI2P-GridCls \cite{li2021deepi2p}, CorrI2P \cite{ren2022corri2p}, GeoLoc \cite{Geoloc} and 2D3D-MATR \cite{li20232d3d}. However, the hyperparameters of the KPConv \cite{thomas2019kpconv} backbone for GeoLoc and 2D3D-MATR can not be generalized across the KITTI Odometry and NuScenes benchmarks. Therefore, we replace the KPConv backbone with the architecture of our CFI2P in Fig. \ref{fig3}, and only adopt their coarse-level supervision in Fig. \ref{fig2} (c) and (d) to learn the features of point sets and pixel patches. This provides a fair comparison to demonstrate the superiority of our proposed quantity-aware correlation. We report the quantitative results in Tab. \ref{tab1}, where the threshold $\boldsymbol{\tau}_d$ is set to 1$\sim$3 pixels for IR and $\boldsymbol{\tau}_m$ is set to 20\% for FMR. By comparing the results of the top four and last rows, it can be seen that the IR and FMR of our method remain optimal under various thresholds on both KITTI Odometry and NuScenes. Notably, in term of IR, our CFI2P backbone combined with the proposed quantity-aware correlation outperforms CorrI2P by 12.4$\sim$26.9\%. It is obvious that our proposed quantity-aware strategy can further improve the matching accuracy according to the results of last three rows. For a more comprehensive depiction, we also display the curves of the metrics across different thresholds in Fig. \ref{fig4}. The visual trends of the curves are consistent with the statistical data.

\begin{figure*}[t]
\centering
\includegraphics[width=1\textwidth]{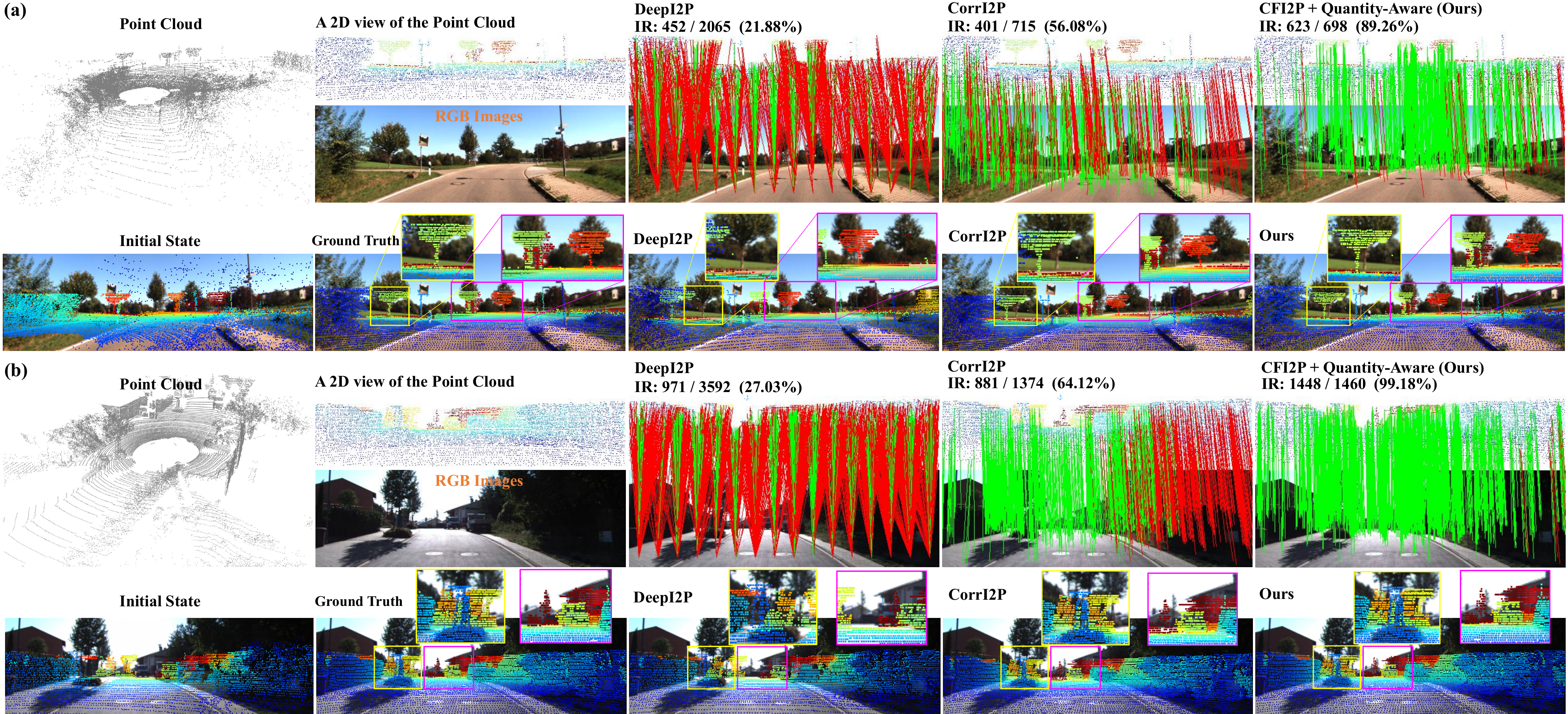} 
\caption{\textbf{Qualitative results on the KITTI Odometry dataset.} To visualize the correct (green) and incorrect (red) correspondences, the point cloud is projected into a narrow 2D view (color represents distance) using the ground-truth camera pose. We show the inlier ratio (correct number / total number) above each sample. To visualize the image-to-point cloud registration performance, we use the predicted camera poses to project the point clouds and fuse them into the RGB images. For DeepI2P, the correspondences are derived using DeepI2P-GridCls, while the registration results are derived using DeepI2P-2D.}
\label{fig7}
\end{figure*}

\begin{figure*}[t]
\centering
\includegraphics[width=1\textwidth]{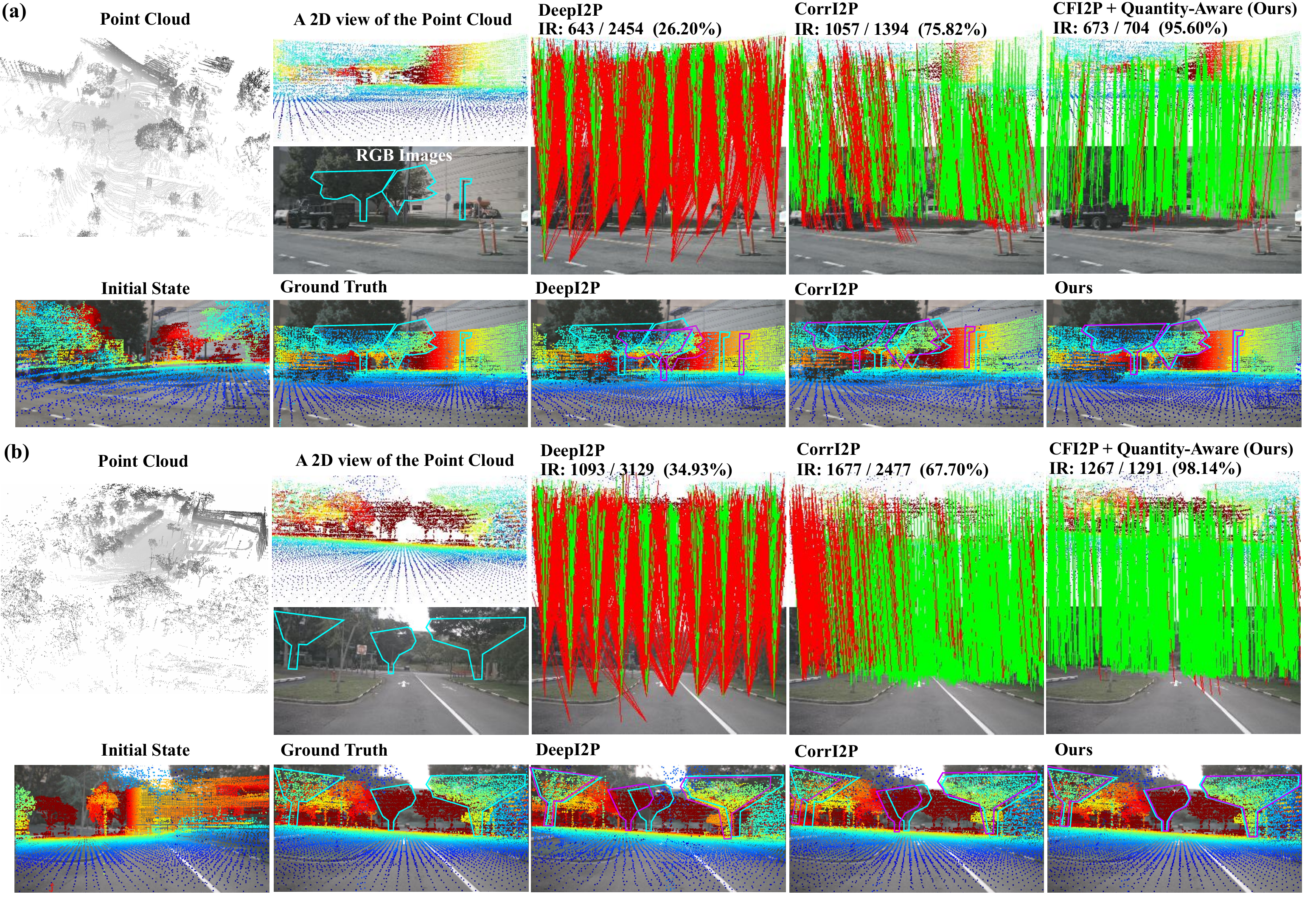} 
\caption{\textbf{Qualitative results on the NuScenes dataset.} To visualize the correct (green) and incorrect (red) correspondences, the point cloud is projected into a narrow 2D view (color represents distance) using the ground-truth camera pose. We show the inlier ratio (correct number / total number) above each sample. To visualize the image-to-point cloud registration performance, we use the predicted camera poses to project the point clouds and fuse them into the RGB images. For clearer comparison, we use \textcolor{cyan}{cyan outlines} to highlight objects in RGB images and \textcolor[RGB]{204,0,255}{purple outlines} for projected 3D objects.}
\label{fig8}
\end{figure*}

\vspace{1mm}
\textbf{Image-to-Point Cloud Registration.} Besides the matching-based algorithms mentioned above, we also add comparisons with some matching-free methods, including DeepI2P-2D\cite{li2021deepi2p}, DeepI2P-3D\cite{li2021deepi2p} and EFGHNet \cite{efghnet}. As illustrated in Tab. \ref{tab2}, it can be seen that our CFI2P framework achieves optimal registration performance by comparing the results of the top five and bottom three rows. Our registration recall (RR) is even above 99\% on both two datasets. Here, the RRE threshold $\boldsymbol{\tau}_r$ is set to $10^{\circ}$ and the RTE threshold $\boldsymbol{\tau}_t$ is set to 5 m. Notably, by comparing the results of the bottom three rows, we see that our CFI2P backbone combined with the proposed quantity-aware correlation also achieves performance improvements. Furthermore, we display the RTE and RRE distributions in Fig. \ref{fig6} (top two). Our result distributions are more concentrated in the regions with smaller errors.
We also show the registration recall curves with various RTE and RRE thresholds in Fig. \ref{fig6} (bottom two). These figures provide a more intuitive depiction of our advantages.

\vspace{1mm}
\textbf{Efficiency.} We report the running times, memory footprint and model parameters in Table. \ref{tab3}. All evaluations are conducted on an NVIDIA RTX 3090 GPU and an Intel i7-12700K CPU.
Although the architecture of CFI2P seems complex, the overall computational and space complexity are not as high as imagined. Most operations (e.g., the self and cross attention) are performed on the proxies or the resampled points and pixels, rather than on original point clouds and images. Compared to DeepI2P with smallest time and memory footprint, the additional 0.037 s and 0.17 GB consumption is worth the performance gain brought by CFI2P. Additionally, we reduce the model parameters by about 67\%.

\begin{table}[t]
\centering
\caption{The efficiency of different frameworks on the KITTI Odometry dataset. The best results are highlighted in bold.}
\renewcommand\arraystretch{1.2}
\begin{tabular}{c|ccc}
\hline \hline
Method & Time (s) & Memory (GB) & Params (MB) \\ \hline
DeepI2P-GridCls \cite{li2021deepi2p}  & \textbf{0.049}   & 2.41       & 100.75     \\ 
DeepI2P-3D \cite{li2021deepi2p}       & 16.584  & \textbf{2.01}       & 100.12     \\ 
DeepI2P-2D \cite{li2021deepi2p}       & 9.381   & \textbf{2.01}       & 100.12     \\ 
EFGHNet \cite{efghnet}       & 0.319   & 2.76       & 574.32     \\ 
CorrI2P \cite{ren2022corri2p}       & 0.088   & 2.88       & 141.07     \\ \hline
CFI2P (Ours)      & 0.084   & 2.18       & \textbf{34.74}      \\ \hline \hline
\end{tabular}
\label{tab3}
\end{table}

\subsection{Qualitative Results}
\textbf{Correspondence and Registration.} We visualize the results on the KITTI Odometry and NuScenes datasets in Fig. \ref{fig7} and \ref{fig8}, respectively. Note that each point cloud in KITTI Odometry is a \textbf{single-frame scan}, whereas each point cloud in NuScenes accumulates the data from \textbf{multiple frame scans}. Therefore, the latter provides a richer and more detailed 3D scene representation. Although our method builds fewer point-to-pixel correspondences, they exhibit the highest quality. More examples are visualized in the multimedia supplementary material. Additionally, the initial state of the projected 3D points is chaotic since the relative camera pose is unknown. The goal of image-to-point cloud registration is equivalent to aligning the corresponding 2D pixels and 3D points of the same object. It is clear that our method has the best visual performance for point-to-pixel alignment, which means that our estimation is closest to the ground truth camera pose.

\begin{figure}[t]
\centering
\includegraphics[width=0.47\textwidth]{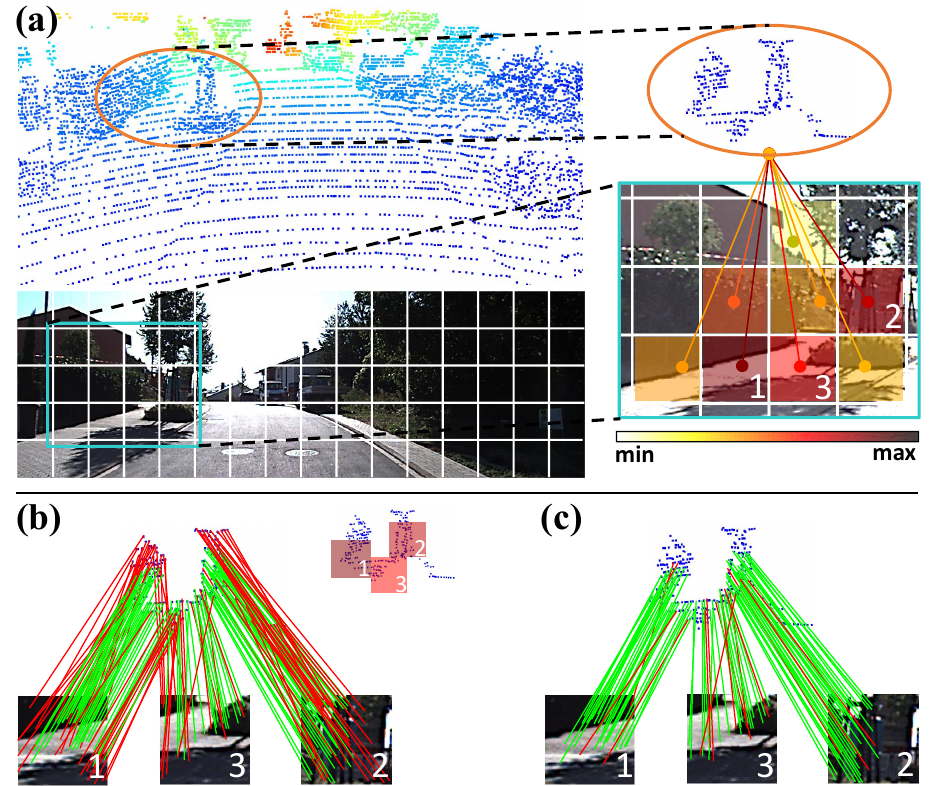} 
\caption{\textbf{Qualitative results from a coarse-to-fine perspective.} (a) The learned quantity-aware correspondences between a point set and several pixel patches. The colors on the patches indicate the degrees of correlation. In default settings, we select the top 3 patches to perform fine matching. (b) Some 3D points, which should be mapped to other pixel patches, are incorrectly mapped into the 3 pixel patches. (c) Our confidence sorting strategy removes many incorrect correspondences according to the learned quantity-aware priors.}
\label{fig9}
\end{figure}

\vspace{1mm}
\textbf{A Coarse-to-Fine Perspective.} We provide an example to illustrate the process of building correspondences. Consistent with the misalignment in Fig. \ref{fig2} (a), the point set in Fig. \ref{fig9} (a) overlaps multiple pixel patches, and our proposed method learns the set-to-patch correspondences with different degrees of correlation. In default settings, we will select the top 3 patches for the subsequent fine matching. Consequently, there are always some 3D points in the point set that should not be mapped into these 3 patches, as shown in Fig. \ref{fig9} (b). This issue is also observed in the previous binary set-to-patch correspondences \cite{Geoloc, li20232d3d}. According to Section \ref{sec:cardiaware}, our supervision for set-to-patch correspondences are aware of the proportion of 3D points in the point set that can be projected into the 3 pixel patches. Based on it, our confidence sorting strategy in Section \ref{sec:fine} can remove these incorrect point-to-pixel correspondences, as illustrated in Fig. \ref{fig9} (c).

\begin{table}[t]
\centering
\caption{Ablation studies of the components in CFI2P.}
\renewcommand\arraystretch{1.2}
\begin{tabular}{ccccc|ccc}
\hline
\hline
AA & CA & C2F & Mask & Sort & IR(\%) & RTE(m) & RRE(°) \\ \hline
           & \checkmark & \checkmark & \checkmark &            & 53.50   & 0.922    &  1.839   \\
\checkmark &            & \checkmark & \checkmark &            & 55.49   & 0.756    &   1.770  \\
\checkmark & \checkmark &            &            &            & 50.83   & 0.907    &  2.252   \\
\checkmark & \checkmark & \checkmark &            &            & 58.02   & 1.254    &  2.275   \\
\checkmark & \checkmark & \checkmark & \checkmark &            & 64.67   & 0.594    &  1.395   \\
\checkmark & \checkmark & \checkmark & \checkmark & \checkmark & \textbf{69.16}   & \textbf{0.541}    &  \textbf{1.383} \\ \hline \hline
\end{tabular}
\label{tab4}
\end{table}

\begin{table}[t]
\centering
\caption{Ablation studies of the coarse-to-fine sampling density.}
\renewcommand\arraystretch{1.2}
\begin{tabular}{cc|ccccc}
\hline \hline
Point($n$) & Pixel($m$)            & IR(\%) & RTE(m) & RRE(°)   & Time(s)    \\ \hline
65    & $1\!\times\!8\!\times\!8$  & 64.11  & 0.702 & 1.765 & 0.069  \\
65    & $2\!\times\!8\!\times\!8$  & 68.11  & 0.583 & 1.501 & 0.076  \\
65    & $3\!\times\!8\!\times\!8$  & 69.16  & 0.541 & 1.383 & 0.084  \\
65    & $4\!\times\!8\!\times\!8$  & 69.01  & 0.536 & 1.376 & 0.095  \\ 
65    & $5\!\times\!8\!\times\!8$  & 69.20  & 0.537 & 1.390 & 0.105  \\ \hline 
25    & $3\!\times\!8\!\times\!8$  & 69.81  & 0.597 & 1.445 & 0.072   \\
35    & $3\!\times\!8\!\times\!8$  & 69.89  & 0.570 & 1.422 & 0.077   \\
45    & $3\!\times\!8\!\times\!8$  & 69.72  & 0.559 & 1.411 & 0.080   \\
55    & $3\!\times\!8\!\times\!8$  & 69.38  & 0.549 & 1.409 & 0.082  \\
\hline \hline
\end{tabular}
\label{tab5}
\end{table}

\subsection{Ablation Studies}
\textbf{Components.} We have conducted ablation studies on the KITTI Odometry dataset to further analyze the effects of different components. The statistics are shown in Tab. \ref{tab3}. We first remove the confidence sorting (\textbf{Sort}) strategy in the fine matching module. The effect of sorting is illustrated in the fifth and last rows.We remove the sampling masks (\textbf{Mask}) embedded in the fine-level interactive learning (i.e., in Eq. \ref{eq10} and Eq. \ref{eq11}) and the fine-level optimal transport module (i.e., in Eq. \ref{eq12}). The effect of the sampling mask is illustrated in the forth and fifth rows. We replace the coarse-to-fine (\textbf{C2F}) operations with the overlap detection and metric learning head in CorrI2P \cite{ren2022corri2p}. In the feature space, we search for the nearest pixel to each point within the predicted overlap region. There is a significant drop in performance when comparing the third and fifth rows. We remove the cross-attention (\textbf{CA}) module in proxy learning, which means that the image and point cloud branches are learned independently. It results in a significant drop in performance when comparing the second and fifth rows. We replace the attentive aggregation (\textbf{AA}) operation with max pooling over the grouped point sets to generate point proxies. The first and fifth rows demonstrate the better local representation ability of this module.

\vspace{1mm}
\textbf{Coarse-to-Fine Sampling Density.} We have investigated the effects coarse-to-fine sampling density. The statistics are reported in Tab \ref{tab5}. For the top five rows, we sample $n\!=\!65$ 3D points in each candidate point set, while gradually increasing the number (i.e., $k$) of sampled pixel patches. Since each pixel patch has $8\!\times\!8$ pixels, the fine matching of each point set is finally performed between 65 3D points and $m\!=\!k\times64$ 2D pixels. It can be seen that when $m$ increases, both the point-to-pixel correspondence and the registration performance are improved, but the trend of improvement diminishes gradually in the later phases. This is because sampling more pixel patches will gradually cover the projection range of each point set, thereby reducing the incorrect correspondences in Fig. \ref{fig9} (b). For the bottom four rows, we gradually increase $n$, while $m$ is kept at 3. Although the change in IR is not obvious, the registration errors still gradually decrease. This is because sampling more 3D points actually generates more point-to-pixel correspondences, which is beneficial for registration. Additionally, the efficiency of our method decreases whether sampling more 3D points or more 2D pixels. It is crucial to find an optimal balance between performance and efficiency.

\begin{figure}[t]
\centering
\includegraphics[width=0.47\textwidth]{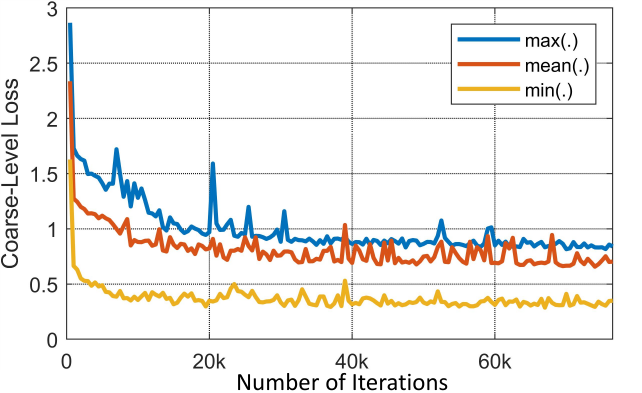} 
\caption{\textbf{The coarse-level loss.} Three aggregation functions are employed to generate the correlation matrices between point sets and pixel patches.}
\label{fig10}
\end{figure}

\vspace{1mm}
\textbf{Aggregation Function $\boldsymbol{\sigma}(\cdot)$.} We have also conducted ablation studies to compare the effects of different aggregation functions $\boldsymbol{\sigma}(\cdot)$ in Eq. \ref{eq5}. We use $\mathrm{min}(\cdot), \mathrm{max}(\cdot)$ and $\mathrm{mean}(\cdot)$ to generate the correlation matrix $\textbf{W}^{c}$, which serves as the weights in the calculation of coarse-level loss. Note that $\textbf{W}^{c}$ will be normalized when calculating the loss in Eq. \ref{eq15}. Thus, the loss increase in Fig. \ref{fig10} is mainly caused by the incorrect correlation measurement.

\section{Conclusion}
\label{sec:conclusion}
In this paper, we propose a novel quantity-aware strategy to quantify the correlations between point sets and pixel patches. Superior to the traditional binary method, our strategy utilizes continuous values to represent the correlation magnitude, which provides richer information to supervise the feature learning. To put it into practice, we present a novel coarse-to-fine correspondence learning framework with a hybrid transformer architecture, named CFI2P. It learns the set-to-patch correspondences using the quantity-aware supervision at the coarse level, and progressively refines the results to point and pixel levels. Particularly, we also propose a confidence sorting strategy to reuse the quantity-aware priors for selecting better point-to-pixel correspondences. Thanks to the reliable correspondences, CFI2P has achieved state-of-the-art performance on the KITTI Odometry and NuScenes benchmarks. 

We hold the firm belief that the proposed approach will not only enhance the current methodology but will also pave the way for more extensive and sophisticated applications, particularly those that require a seamless integration of both point cloud and image data. The versatility and adaptability of our method are expected to contribute significantly to the advancement of various fields, enabling richer applications and unlocking novel possibilities in domains such as computer vision, robotics, and augmented reality.

\bibliographystyle{IEEEtran}
\bibliography{reference}

\end{document}